\documentclass[letterpaper, 10 pt, conference]{ieeeconf}

\IEEEoverridecommandlockouts 
\usepackage{cite}
\usepackage{amsmath,amssymb,amsfonts}
\usepackage{graphicx}
\usepackage{textcomp}
\usepackage{bm}
\usepackage{float}
\usepackage{color}
\usepackage{tikz}
\usepackage{subcaption}
\usepackage{multicol}
\DeclareMathOperator*{\argmin}{arg\,min}
\DeclareMathOperator*{\argmax}{arg\,max}
\usepackage[hidelinks]{hyperref}
\usepackage[T1]{fontenc}
\usepackage{algorithm}
\usepackage[noend]{algpseudocode} %

\graphicspath{ {./figures/} }
\begin{document}

\title{Meta-Reinforcement Learning Using Model Parameters}
\author{Gabriel Hartmann$^{1,2}$ and Amos Azaria$^{2}$%
\thanks{This research was supported, in part, by the Ministry of Science \& Technology, Israel.}
\thanks{$^{1}$ Department of Mechanical Engineering and Mechatronics, 
Ariel University, Israel}
\thanks{$^{2}$ Department of Computer Science, 
Ariel University, Israel}
\thanks{{\tt\footnotesize 
gabrielh@ariel.ac.il, amos.azaria@ariel.ac.il}}
}

\maketitle

\begin{abstract}
In meta-reinforcement learning, an agent is trained in multiple different environments and attempts to learn a meta-policy that can efficiently adapt to a new environment. %
This paper presents RAMP, a \textit{ Reinforcement learning Agent using Model Parameters} that utilizes the idea that a neural network trained to predict environment dynamics encapsulates the environment information. %
RAMP is constructed in two phases: in the first phase, a multi-environment parameterized dynamic model is learned. In the second phase, the model parameters of the dynamic model are used as context for the multi-environment policy of the model-free reinforcement learning agent. 
We show the performance of our novel method in simulated experiments and compare them to existing methods.
\end{abstract}

\section{Introduction}
Common approaches for developing controllers do not rely on machine learning. Instead, engineers manually construct the controller based on general information about the world and the problem. After repetitively testing the controller in the environment, the engineer improves the controller based on the feedback from these tests. That is, a human is an essential part of this iterative process.
Reinforcement Learning (RL) reduces human effort by automatically learning from interaction with the environment. Instead of explicitly designing and improving a controller, the engineer develops a general RL agent that learns to improve the controller's performance without human intervention. 
The RL agent is usually general and does not include specific information about the target environment; this allows it to adapt to different environments. Indeed, RL agents may achieve higher performance compared to human-crafted controllers \cite{mnih2015human,silver2018general,GranTurismo}. %
However, RL agents usually require training from the ground up for every new environment, which requires extensive interaction in the new environment. %

One solution to speed up the training time is to explicitly provide human-crafted information about the environment (context) to the RL agent \cite{hartmann2019deep}. However, such a solution requires explicitly analyzing the target environment, which may be challenging and time-consuming. 

Instead of relying on the human understanding of the problem for providing such context, a \textit{meta-Reinforcement Learning} (meta-RL) agent can learn to extract a proper environmental context. To that end, a meta-RL agent is trained on extended interaction in multiple different environments, and then, after a short interaction in a new, unseen environment, it is required to perform well in it \cite{maml,RL2}. 
Specifically, a meta-RL algorithm that is based on context extraction is composed of two phases. First, in the meta-learning phase, the agent learns a general policy suitable to all environments given a context. Additionally, in this phase, the meta-RL agent learns how to extract a context from samples obtained from an environment. Secondly, in the adaptation phase, the meta-RL agent conducts a short interaction in the new environment, and the context is extracted from it. This context is then fed to the general policy, which acts in the new environment.

One common approach for context extraction is using a Recurrent Neural Network (RNN). That is, the RNN receives the history of the states, actions, and rewards and is trained to output a context that is useful for the general policy. However, the RNN long-term memory capability usually limits the effective history length \cite{graves2014neural}. 
Additionally, since the context vector is not explicitly explainable, it is difficult to examine the learning process and understand if the RNN learned to extract the representative properties of the environments.

In this paper, we introduce RAMP -- a Reinforcement learning Agent using Model Parameters. We utilize the idea that a neural network trained to predict environment dynamics encapsulates the environment properties; therefore, its parameters can be used as the context for the policy.
During the meta-RL phase, RAMP learns a neural network that predicts the environment dynamic for each environment. However, since the number of the neural network's parameters is usually high, it is challenging for the policy to use the entire set of parameters as its context.  
Therefore, the majority of the model's parameters are shared between all environments, and only a small set of parameters are trained separately in each environment. In that way, the environment-specific parameters represent the specific environment properties. 
Consequently, a general policy uses only these parameters as context and outputs actions that are suitable for that particular environment.
One advantage of RAMP is that the history length used for the context extraction is not limited because the context is extracted from a global dynamic model. Additionally, the combination of model learning and RL in RAMP makes the training process more transparent since it is possible to evaluate the performance of the model learning process independently. 
We demonstrate the effectiveness of RAMP in several simulated experiments in Sec. \ref{sec:Experimental Evaluation}.

To summarize, the contributions of this paper are:
\begin{itemize}
    \item Suggesting a novel method for meta-reinforcement learning. 
    \item Presenting a multi-environment dynamic model learning method that adapts to new environments by updating only a few parameters.
    \item Using the dynamic model parameters directly as a context for the general policy.
    \item Combining model-based and model-free RL.
\end{itemize}

\section{Related Work}
RL has shown success in numerous domains, such as playing Atari games \cite{mnih2013playing,mnih2015human}, playing Go \cite{silver2016mastering}, and driving autonomous vehicles \cite{GranTurismo,survey_driving}.
Some are designed for one specific environment \cite{silver2017mastering,hsu2002behind}, while others can learn to master multiple environments \cite{silver2018general,mnih2013playing}; however, many algorithms require separate training for each environment. 

Several approaches were proposed to mitigate the need for long training times by using meta-RL methods. %
We begin by describing methods that, similarly to ours, learn a context-conditioned, general policy. However, they constructed the context vector in different ways.
We note that some previous works term the different training environments ``tasks" since they emphasize the changes in the reward function. However, since our work focuses on environments with different dynamics (transition functions), we use the term ``environments''.
In \cite{yu2017preparing}, the environment properties are predicted by a neural network based on a fixed, small number of steps. However, this approach requires explicitly defining the representative environment properties. Moreover, it assumes that these properties can be estimated based on the immediate environmental dynamics. 
Rasool et al. \cite{mql} introduce TD3-context, a TD3-based RL agent that uses a recurrent neural network (RNN) to create a context vector, which receives the recent states and rewards as input. 
However, even though types of RNNs such as LSTM \cite{LSTM} and GRU \cite{GRU} are designed for long-term history, in practice, the number of previous states considered by the RNN is limited \cite{graves2014neural}. Therefore, if an event that defines an environment occurs too early, the RNN will ``forget" it and not provide an accurate context to the policy.
In our method, RAMP, the context consists of the parameters of a global, dynamic model, which is not limited by the history length.  
Other approaches use the RNN directly as a policy, based on the transitions and rewards during the previous episode \cite{RL2,Learning_to_reinforcement_learn}, instead of creating a context vector for a general policy. These approaches are also vulnerable to this RNN memory limitation.

Finn et al. \cite{maml} proposed a different principle for meta-learning termed ``Model-Agnostic Meta-Learning (MAML)." In MAML, the neural network parameters are trained such that the model will be adapted to a new environment %
by updating all parameters only with a low number of gradient-descent steps. %
However, the training process of MAML may be challenging \cite{How_to_train}. Furthermore, MAML uses on-policy RL and therefore is unsuitable for the more sampling-efficient off-policy methods as in our approach. 
Nevertheless, since MAML can also be used for regression, we compare our multi-environment dynamic model learning method to MAML in Sec. \ref{sec:Sine Waves Regression}.

Some proposed meta-RL methods are suitable for off-policy learning \cite{pearl,mql}.
Meta-Q-learning (MQL) \cite{mql} updates the policy to new environments by using data from multiple previous environments stored in the replay buffer. The transitions from the replay buffer are reweighed to match the current environment. %
We compare our method, RAMP, to MQL in our testing environment in Sec. \ref{sec:multi environment RL}.

As opposed to all these meta-RL methods, which are model-free, also model-based meta-RL methods were proposed. In model-based meta-RL, the agent learns a model that can quickly adapt to the dynamics of a new environment. 
Ignasi et al. \cite{mb_meta_rl} propose to use recurrence-based or gradient-based (MAML) online adaptation for learning the model. Similarly, Lee et al. \cite{pmlr-v119-lee20g} train a model that is conditioned on the encoded, previous transitions.
In contrast to model-free RL, which learns a direct mapping (i.e., a policy) between the state and actions, model-based RL computes the actions by planning (using a model-predictive controller) based on the learned model. In our work, we combine the model-free and model-based approaches resulting in rapid learning of the environment dynamic model and a direct policy without the need for planning.

\section{Problem Definition}
We consider a set of $N$ environments that are modeled as a Markov Decision Processes $\mathcal{M}^k = \{\mathcal{S},\mathcal{A},\mathcal{T}^k,\mathcal{R}\}$, $k=\{1,\dots, N\}$. All environments share the same state space $\mathcal{S}$, action space $\mathcal{A}$, and reward function $\mathcal{R}$ and differ only by their unknown transition function $\mathcal{T}$.  
These $N$ environments are randomly split into training environments ${\mathcal{M}}_{\text{train}}$ and testing environments $\mathcal{M}_{test}$.

The meta-RL agent is trained on the $\mathcal{M}_{\text{train}}$ environments and must adapt separately to each of the $\mathcal{M}_{\text{test}}$ environments.
That is, the agent is permitted to interact with the $\mathcal{M}_{\text{train}}$ environments for an unlimited number of episodes. Then, the meta-RL agent is given only a short opportunity to interact with each of the $\mathcal{M}_{\text{test}}$ environments (e.g., a single episode, a number of time steps, etc.), and update its policy based on this interaction. Overall, the agent's goal is to maximize the average expected discounted for each of the $\mathcal{M}_{\text{test}}$ environments.

\section{RAMP} %
RAMP is constructed in two phases: in the first phase, a multi-environment dynamic model is learned, and in the second phase, the model parameters of the dynamic model are used as context for the multi-environment policy of the reinforcement learning agent.  
The following sections first describe how the multi-environment dynamic model is learned by exploiting the environments' common structure. In the second part, we describe the reinforcement learning agent. %

\subsection{Multi-Environment Dynamic Model}
Attempting to approximate the transition function of each environment $\mathcal{T}^k$ by an individual neural network is likely to work well for the training environments. However, it is unlikely to generalize to the testing environments, as we have only a limited set of data points for them.
However, since the environments share a common structure, it will be more efficient to train a neural network that has shared components between all the environments. 
Namely, we intend to train a general neural network based on the training environments such that it can be adapted to each testing environment using only a limited set of data points.

In addition, since RAMP's second phase uses the neural network's parameters' values directly as a context for the RL agent, we wish to use only a small number of parameters that should represent the properties of each specific environment dynamics. 
Therefore, the general neural network shares the vast part of the parameters between all environments and includes only a small set of environment-specific parameters. %
The environment-specific parameters are, in fact, a compact representation of each environment; therefore, they can be used by RAMP as a context vector (as described in Sec. \ref{sec:RL}).

We approximate the transition function of all environments by a neural network with parameters indexed by $\varphi$, which are split to environment-specific parameters indexes $\omega \subseteq \varphi$, and to the remaining parameters indexes $\sigma = \varphi \setminus \omega$. The values of the parameters of each environment $k$ are denoted by $\hat{\varphi}^k$ and the environment-specific parameters' values by $\hat{\omega}^k$. The shared parameters' values, which do not depend on a specific environment, are denoted by $\hat{\sigma}$. 
Our multi-environment dynamic model is denoted by  $f_{\hat{\sigma},\hat{\omega}_k}$. %
The multi-environment dynamic model is given a state $s$ and action $a$ and outputs a prediction of the state at the following time step $s'$ for each environment $\mathcal{M}^k $, i.e.,
$s' = f_{\hat{\sigma},\hat{\omega}_k}(s,a)$.

We now describe how to select $\omega$ and how to train the neural network parameters $\hat{\sigma}$ and $\hat{\omega}^k$. At first, we gather sufficient data in the form of
$D^k = \{(s,a,s')\}$, for each environment $k$.
At the beginning of the training process $\omega = \varnothing$ and $\sigma = \varphi$. The network is trained using the gradient descent algorithm to minimize the loss, which is the squared error between the predicted and real next state for each environment: 
\begin{equation}
\mathcal{L}(D^k,\hat{\sigma},\hat{\omega}^k) =  \sum\limits_{s,a,s' \in D^k} (s' - f_{\hat{\sigma},\hat{\omega}^k}(s,a))^2 .
\end{equation}
Initially, $\hat{\sigma}$ are trained in all environments to achieve an average model prediction:
\begin{equation}
\hat{\sigma} = \argmin_{\hat{\sigma}}{\sum\limits_{k=1}^{|\mathcal{M}_{\text{train}}|} \mathcal{L}(D^k,\hat{\sigma},\hat{\omega}^k)}  
\end{equation}

After the initial training phase, the parameters $\omega$ are selected from $\varphi$ by the algorithm, one-at-a-time.
Intuitively, the algorithm should select parameters for $\omega$ that have the greatest impact on the difference between the environments. Therefore, at each gradient step, the gradient of the loss function $\mathcal{L}(D^k,\hat{\sigma},\hat{\omega}^k)$ relative to $\hat{\varphi}^k$ is computed for each environment $k$: 
\begin{equation}
g^k = \nabla_{\hat{\varphi}^k}\mathcal{L}(D^k,\hat{\sigma},\hat{\omega}^k), \end{equation}
and the parameter with the highest variance between all gradients $g^k$ is added to $\omega$: 
\begin{equation}
\omega \leftarrow \omega \cup \argmax_{i \in \varphi  \setminus \omega} \space \text{var}({g^0_i,g^1_i,\dots, g_i^{|\mathcal{M}_{\text{train}}|}}).
\end{equation}
Then, the network is trained to minimize the loss function in all environments:
\begin{equation}
\min_{\hat{\sigma}}{\sum\limits_{k=1}^{|\mathcal{M}_{\text{train}}|} \min_{\hat{\omega}^k} \mathcal{L}(D^k,\hat{\sigma},\hat{\omega}^k)}.
\end{equation}
That is achieved by updating the environment-specific parameters $\hat{\omega}^k$ by the corresponding gradient:
\begin{equation}
\hat{\omega}^k \gets \hat{\omega}^k - \alpha_{\omega} g^k,
\end{equation}
and updating the shared parameters by the average gradient:
\begin{equation}
\hat{\sigma} \gets \hat{\sigma} - \alpha_{\sigma}\frac{1}{|\mathcal{M}_{\text{train}}|}\sum_{i=0}^{|\mathcal{M}_{\text{train}}|}{g^i}, 
\end{equation}
where $\alpha_{\omega}$ and $\alpha_{\sigma}$ are the learning rates.
During the training, parameters continue to be added to $|\omega|$ until it reaches a predefined size $n_{\omega}$. Algorithm \ref{alg:model_learning} summarizes the multi-environment dynamic model learning.

Finally, at the end of the training process (after achieving a low loss value), only parameters $\omega$ need to be adjusted for a new environment to get an accurate dynamic model, while parameters $\sigma$ remain constant. That is,
\begin{equation}
   \hat{\omega}^k =  \argmin_{\hat{\omega}^k} \mathcal{L}(D^k,\hat{\sigma},\hat{\omega}^k).
\end{equation}

\begin{algorithm}
\caption{Model learning with RAMP}\label{alg:model_learning}
\begin{algorithmic}
\Require $n_{\omega}$ \Comment{Number of environment-specific parameters}
\Require $n_{\text{init}}$ \Comment{Number of steps for initial training}
\Require $n_{\text{tot}}$ \Comment{Number of total training steps}
\Require $\alpha_{\omega},\alpha_{\sigma}$ \Comment{Learning rates}
\Require $\{{D}^0,\dots, {D}^{|\mathcal{M}_{\text{train}}|}\}$ \Comment{Data from the environments}

\State $\omega \gets \varnothing$ 
\For{$i\gets 1$,number of training steps}  %

\For {$\mathcal{M}^k \in \mathcal{M}_{\text{train}}$}
\State sample a batch of transitions $b^k \in {D}^k$
\State $g^k \gets \nabla_{\hat{\varphi}^k}\mathcal{L}(b^k,\hat{\sigma},\hat{\omega}^k)$
\State $\hat{\omega}^k \gets \hat{\omega}^k - \alpha_{\omega} g^k$
\EndFor
\State $\hat{\sigma} \gets \hat{\sigma} - \alpha_{\sigma}\cdot \text{avg}(g^0,\dots, g^{|\mathcal{M}_{\text{train}}|})$

\If {$|\omega| \leq  n_{\omega}$ and $i > n_{\text{init}}$} 
\State $\omega \leftarrow \omega \cup \argmax_{i \in \varphi  \setminus \omega} \space \text{var}({g^0_i,g^1_i,\dots, g_i^{|\mathcal{M}_{\text{train}}|}})$
\EndIf

\EndFor
\end{algorithmic}
\end{algorithm}

\subsection{Reinforcement Learning With Model Parameters Context}
\label{sec:RL}
The multi-environment dynamic model parameters, described in the previous section, are used as a context for the RL agent.
That is, RAMP concatenates the environment-specific parameters' values $\hat{\omega}^k$ to the state $s$ for training the RL agent.

Unfortunately, the environment-specific parameters $\hat{\omega}^k$ do not necessarily converge to the same value when trained in the same environment since the amount of these parameters may be greater than the degree of freedom between the environments. That is, there may be more than one way (i.e., single parameters' values) to minimize the multi-environment dynamic model network. Therefore, the RL agent should be trained on multiple possible representations of each environment. To achieve this,  the environment-specific parameters' values $\hat{\omega}^k$ are retrained every $H$ episodes for each environment $k$ by collecting data from a single episode with the current policy. These values are stored in $\Omega^k$. 
The shared parameters, $\hat{\sigma}$, remain constant during the entire RL multi-environment training phase.

If the RL algorithm uses a replay buffer, %
If the RL algorithm uses a replay buffer, the current environment index $k$ is added to each tuple in addition to the standard data stored in the replay buffer (i.e., state, action, next state, reward, and done).
When sampling from the replay buffer, a context vector is concatenated to each state according to the tuple's environment index $k$. That context vector, which is the values of the environment-specific parameters $\hat{\omega}^k$, is randomly sampled from $\Omega^k$.

We note that RAMP can be used with any RL algorithm and also supports off-policy algorithms, which are considered to be more efficient. 
In this work, we use TD3 \cite{fujimoto2018addressing}, which is an off-policy, actor-critic RL algorithm.
The TD3 agent contains critic neural networks that estimate the action-value function. The critic is trained by minimizing the Bellman function.
The actor, which is a policy represented by a neural network, aims to maximize the expected discounted infinite episode reward by maximizing the action-value function. 
RAMP using the TD3 algorithm is summarized in \ref{alg:RL_with_context}.

\begin{algorithm}
\caption{RAMP (using TD3)}
\label{alg:RL_with_context}
\begin{algorithmic}
\Require  $n_{\text{tot}}$ \Comment{Number training steps}
\Require  $\mathcal{M}_{\text{train}}$ \Comment{Training environments}
\Require $\hat{\omega}^{k}, k=\{1,|\mathcal{M}_{\text{train}}|\}$
\State Initialize critic, actor, and replay buffer $\mathbf{B}$
\State Add $\hat{\omega}^{k}$ to $\Omega^k$ for all $k=\{1,|\mathcal{M}_{\text{train}}|\}$
\While{$i <n_{\text{tot}}$}

\State Select environment $\mathcal{M}^k$ from $\mathcal{M}_{\text{train}}$ randomly
\State Select parameters $\hat{\omega}^k$ from $\Omega^k$ randomly
\While{not done} %
\State Observe $s$ 
\State Execute action $a = \pi_\Phi(s,\hat{\omega}^k)+\epsilon)$, $\epsilon \sim \mathcal{N}(0,\sigma)$ %

\State  Observe new state $s'$, reward $r$ and done flag $d$
\State Add $(s,a,s',r,d,k)$ to replay buffer $\mathbf{B}$

\If{$i$ mod $H = 0$}
\ForAll{$\mathcal{M}^k \in \mathcal{M}_{\text{train}}$}
\State Select random parameters $\hat{\omega}^k$ from $\Omega^k$
\State Sample one episode from $\mathcal{M}^k$:
\State ${D^k \gets \{(s_t,\pi(s_t,\hat{\omega}^k),s_{t+1})\}}_{t=\{1,T\}}$
\State Retrain $\hat{\omega}^{k}_{\text{new}}$ with $D^k$ and add to $\Omega^k$
\EndFor
\EndIf

\State Sample random batch $b  \subset \mathbf{B}$

\ForAll{$(s,a,s',r,d,k) \in  b$}
\State Sample $\hat{\omega}^k$ from $\Omega^k$ 
\State Set $s' \gets (s',\hat{\omega}^k)$
\State Set $s \gets (s,\hat{\omega}^k)$
\EndFor
\State Update critics using $b$  
\If{$i$ mod $n_{\text{training}} = 0$}
\State Update actor using $b$  
\EndIf
\State $i \gets i+1$
\EndWhile
\EndWhile
\end{algorithmic}
\end{algorithm}

\section{Experimental Evaluation}
\label{sec:Experimental Evaluation}
We evaluate RAMP on two domains. The first domain, a sine waves regression test, evaluates the first phase of RAMP alone, i.e., the multi-environment dynamic model learning algorithm. The second domain is the vehicle target-reaching domain, in which vehicles with different dynamics aim to reach a target. The vehicle target-reaching domain tests the complete RAMP algorithm, composed of both phases.

\subsection{Sine Waves Regression}
\label{sec:Sine Waves Regression}
We used a sine waves regression test similar to \cite{maml}. The multi-environment dynamic model was trained on random samples of a sine wave function with different amplitudes $A$ and phases $\phi$:
\begin{equation}
y = A\sin{(x+\phi)}. 
\end{equation} 
The input to the function, $x$, is sampled uniformly from the range $[-5,5]$. The amplitudes of the different functions, are sampled from $A \in [0.1,5]$, and the phases are sampled from $\phi \in [0,\pi]$.
The network consists of two fully connected hidden layers, with $40$ neurons in each layer and ReLU activation. The size of the environment-specific parameters is limited to $10$, i.e. $n_\omega = 10$, out of a total $1761$ parameters.
Contrary to the dynamic model prediction, which receives an action in addition to the current state to predict the next state, in this simple sine regression problem, there is a single input and a single output.
The multi-environment model was trained on $100$ random sine waves with $10$ samples each. %
It was then retrained by updating only environment-specific parameters, $\omega$, %
on $10$ samples of new sine waves. We compare the multi-environment dynamic model of RAMP to a small network composed of only $10$ parameters trained on each new sine wave separately and to MAML \cite{maml}, which updates the entire network ($1761$ parameters). %

\begin{figure}[ht]
    \centering
    \includegraphics[width=\linewidth]{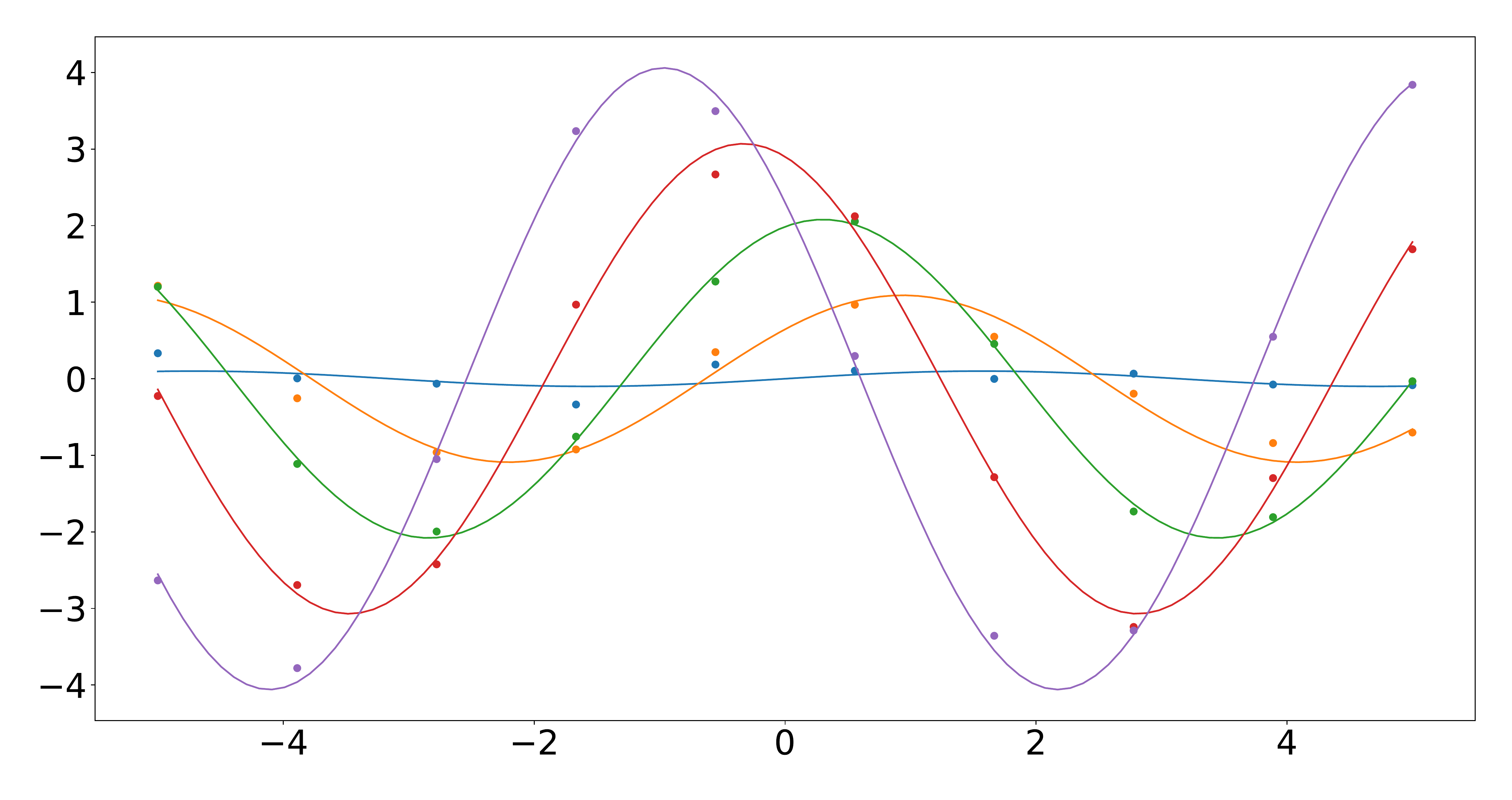}
    \caption{Evaluation of sine functions with different amplitudes and phases. The solid lines represent the ground-truth functions, and the dots are the predictions.} %
    \label{fig:sin_retrain}
\end{figure}

The multi-environment dynamic model achieved a Mean Squared Error (MSE) of $0.021$. 
This result is slightly lower than MAML, which achieved an MSE of $0.037$. 
Nevertheless, since MAML uses $1761$ parameters, it is impractical to use them as a context for the RL agent.
As expected, the network that contains only $10$ parameters resulted in a very high average MSE, $19.25$.
When training on all sine waves together (i.e., all model's parameters are shared without environment-specific parameters), the MSE was $1.9$.
Figure \ref{fig:sin_retrain} depicts the performance of the multi-environment dynamic model of RAMP on the test set.

\subsection{Vehicle Target-Reaching Domain}
The vehicle target-reaching domain is a simple domain that enables us to provide a precise analysis of RAMP's behavior and demonstrate the concepts behind RAMP.
In this domain, an agent controls the vehicle's throttle and brake and aims to reach a target line in a minimum time. The vehicle must reach the target line at a speed of at most $v_{\text{max}}$.
The state, $s = \{v,d\}$, consists of the current vehicle's speed, $v$, and the distance to the target $d$. $v$ ranges from $0$ to $30$ m/s, the distance to the target at the beginning of the episode is $d=40$ m, and the desired maximal speed at the target line $v_{\text{max}}$ is $5$ m/s. The continuous throttle/brake action, $a$ ranges from $-1$ to $1$. The sampling frequency is $25$ Hz. 
The reward function returns $-0.002$ at each non-terminal step. When approaching the target line with a higher speed, than $v_{\text{max}}$ the reward is $0.01(v - v_{\text{max}})^2$ otherwise $0$.

We construct $24$ vehicle target-reaching environments, split into $22$ for training the multi-environment model and two for testing. All vehicles from the different environments have identical acceleration but a different deceleration, which is unknown to the agent. Specifically, the throttle command $a=[0,1]$ causes an acceleration value $\dot{v} =[0,42] $ m/s\textsuperscript{2} in all environments. However, the brake command $a=[-1,0)$ causes a deceleration value $\dot{v} =[0,42k_a]$ that is scaled down by the braking factor $k_a$, which has a value between $0.1$ and $1$.
The braking factor in the test environment is $k_a = 0.925$ and $k_a = 0.175$, which are close to the extremes of all factors.

We begin by evaluating the performance of the multi-environment dynamic model learning process, and then we evaluate the performance of the RL learning procedure.
Finally, we show the adaptation process in a new environment.

\subsubsection{Multi-Environment Dynamic Model Learning}
\begin{figure}[ht]
    \centering
    \includegraphics[width=\linewidth]{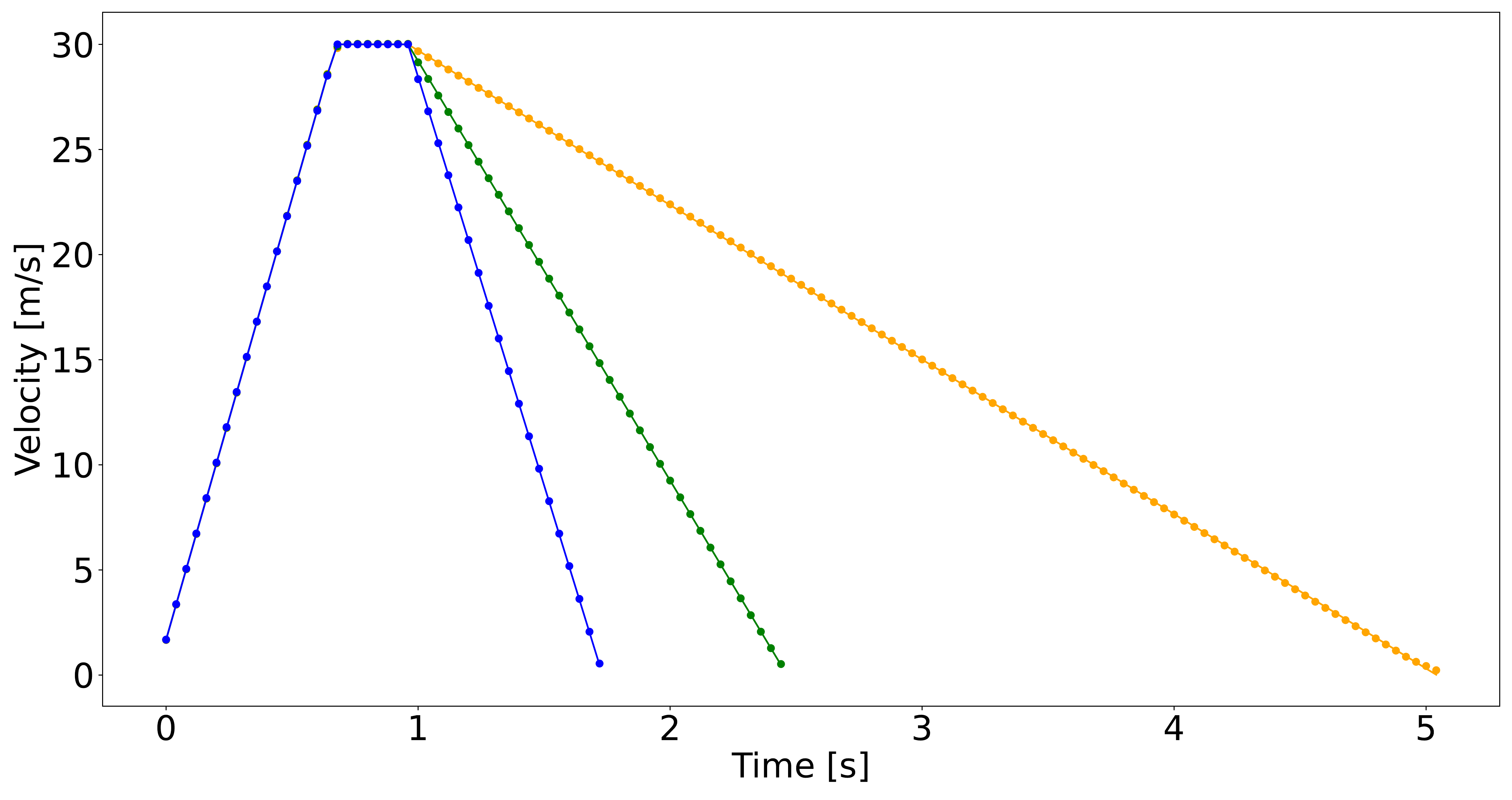}%
    \caption{Speeds prediction of different vehicles. Points are the predictions, and solid lines are the ground-truth speeds.}
    \label{fig:velocities}
\end{figure}

The multi-environment neural network is identical to the network used for the sine wave regression. 
The dynamic model state consists only of the vehicle's speed, and the network predicts the difference between the current and new states.
In each environment, $100$ points are randomly sampled for training, and only $10$ points are sampled from new environments for the adaptation process.
Figure \ref{fig:velocities} shows speeds of $3$ different vehicles. All accelerate at the same rate until reaching the maximal speed, and then, each vehicle applies a maximal braking action ($a = -1$), resulting in different deceleration values. The points represent the predicted speeds, and the solid lines are the real speeds.  
Our multi-environment dynamic model results in an MSE of $0.00029$ on new environments compared to an MSE of $0.045$ when trained on all environments together.

\begin{figure}[ht]
    \centering
    \includegraphics[width=\linewidth]{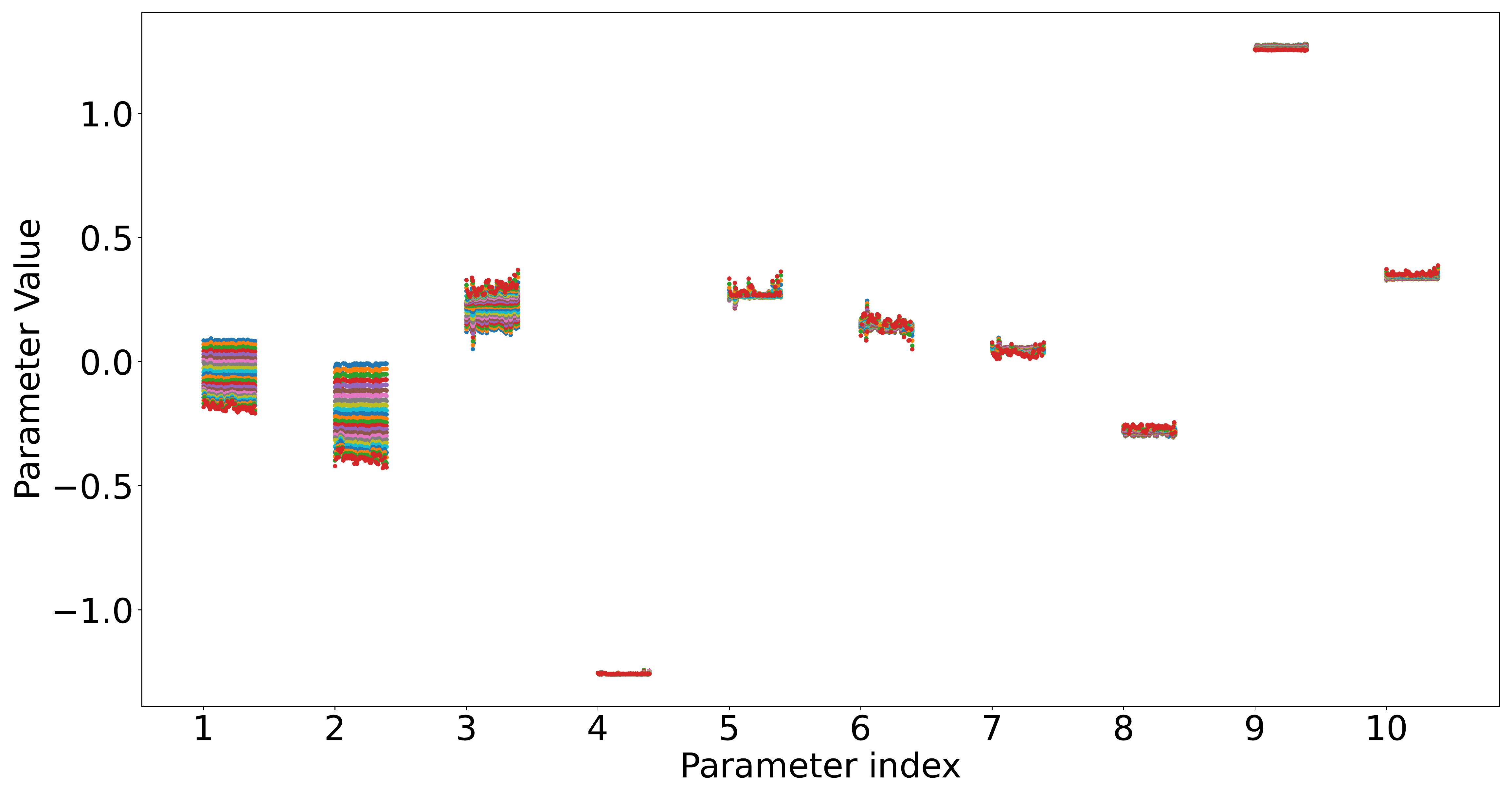}
    \caption{Values of the $10$ environment-specific parameters. Different colors represent different environments. Multiple parameter sets are shown with a shift along the horizontal axis.}
    \label{fig:parameter_values}
\end{figure}

Recall that the environment-specific parameters $\omega$  are retrained multiple times during the RL training process as described in Sec. \ref{sec:RL}. %
Figure \ref{fig:parameter_values} shows the values of each of the $10$ parameters for every environment, in different colors. The different parameter sets are slightly shifted along the horizontal axis. As depicted by the figure, the environment-specific parameters converged to similar values; this can be seen by the consistency of the values between the parameter sets. In addition, the figure shows that the different environments result in noticeable, different values for the first three parameters. In contrast, the remaining parameters show only a minor variance between the environments. This result seems reasonable, since not all parameters are required to determine the variance of the vehicle dynamics, which in fact, has only one degree of freedom. 

\begin{figure}[ht]
    \centering
    \includegraphics[width=\linewidth]{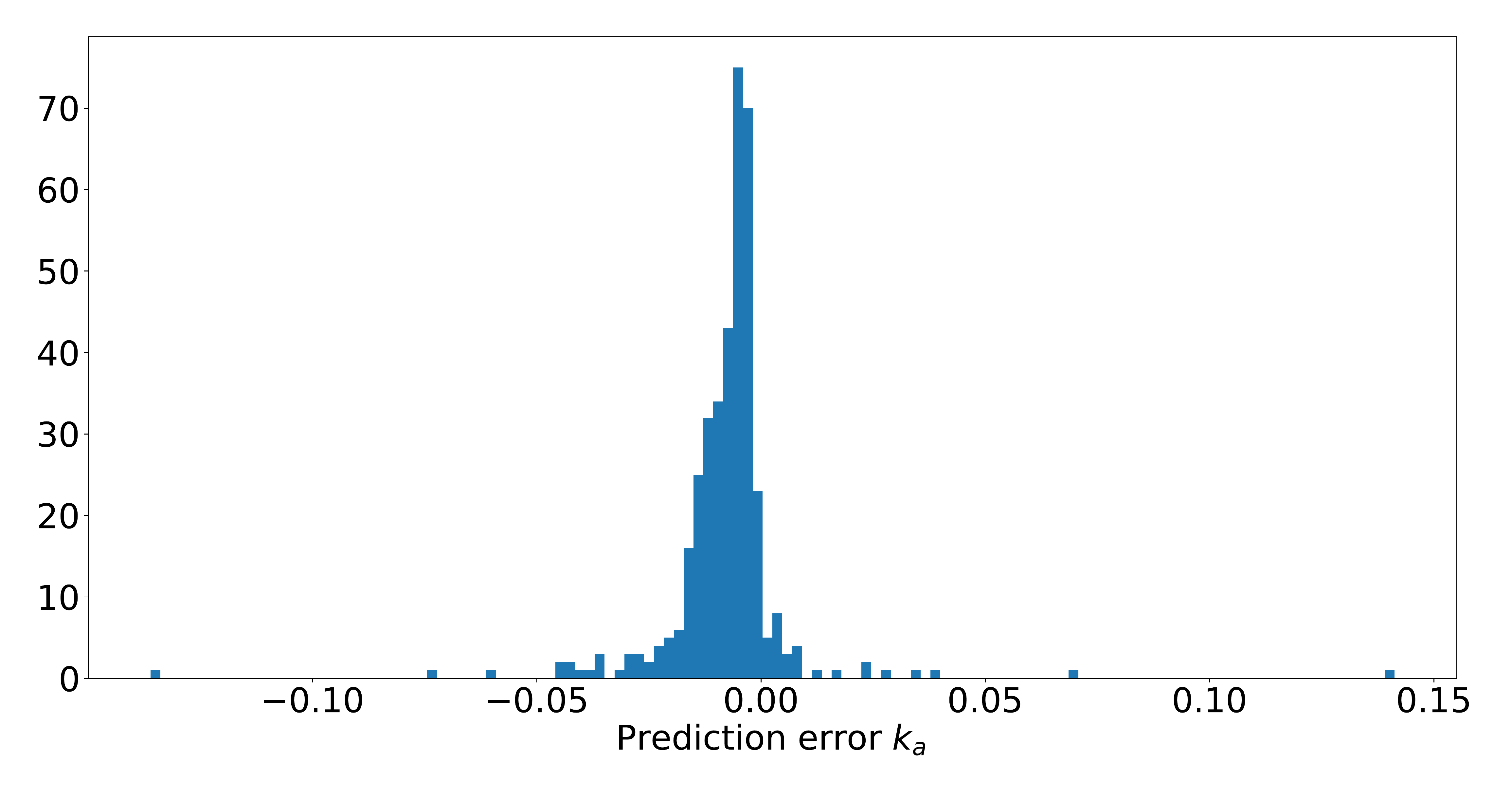}%
    \caption{Error distribution of predicting the real braking factor of environment $k$, based only on the environment-specific parameters $\hat{\omega}^k$.}
    \label{fig:regression_from_params}
\end{figure}

Recall that in the RL training phase, the general policy must extract the properties of the environments from only the environment-specific parameters. Therefore, beyond the low loss of the prediction, we tested that it is possible to directly predict the braking factor from the environment-specific parameters. 
To that end, we trained a dedicated regressor, which is not used by RAMP, on $58$ sets of the trained environment-specific parameters created during the RL training process. 
$40$ environments were used as a training set, and the $18$ remaining environments were used as a test set.
The regressor is composed of a neural network with two hidden layers with $100$ neurons each.
Figure \ref{fig:regression_from_params} shows the prediction error distribution of the braking factor. 
As depicted by the figure, the regressor predicts the braking factor, which ranges from $0.1$ to $1.0$, with an average error of $-0.0077$.

\subsubsection{Multi-Environment Reinforcement Learning}
\label{sec:multi environment RL}
We compared RAMP to the following $3$ other RL agents. The Oracle RL receives explicit information about the vehicle; that is, the braking factor is added to the state. With full knowledge of the environmental properties, the Oracle RL is expected to find a nearly optimal solution. Next, we consider a basic RL that is trained in all environments together, without any identification input, and thus cannot distinguish between different vehicles. Therefore, it is expected to learn a conservative policy that enables safe deceleration to the target line, even for the vehicle with the lowest braking capability. The third RL agent is the meta-Q-learning (MQL) algorithm \cite{mql}.

\begin{figure}[ht]
    \centering
    \includegraphics[width=\linewidth]{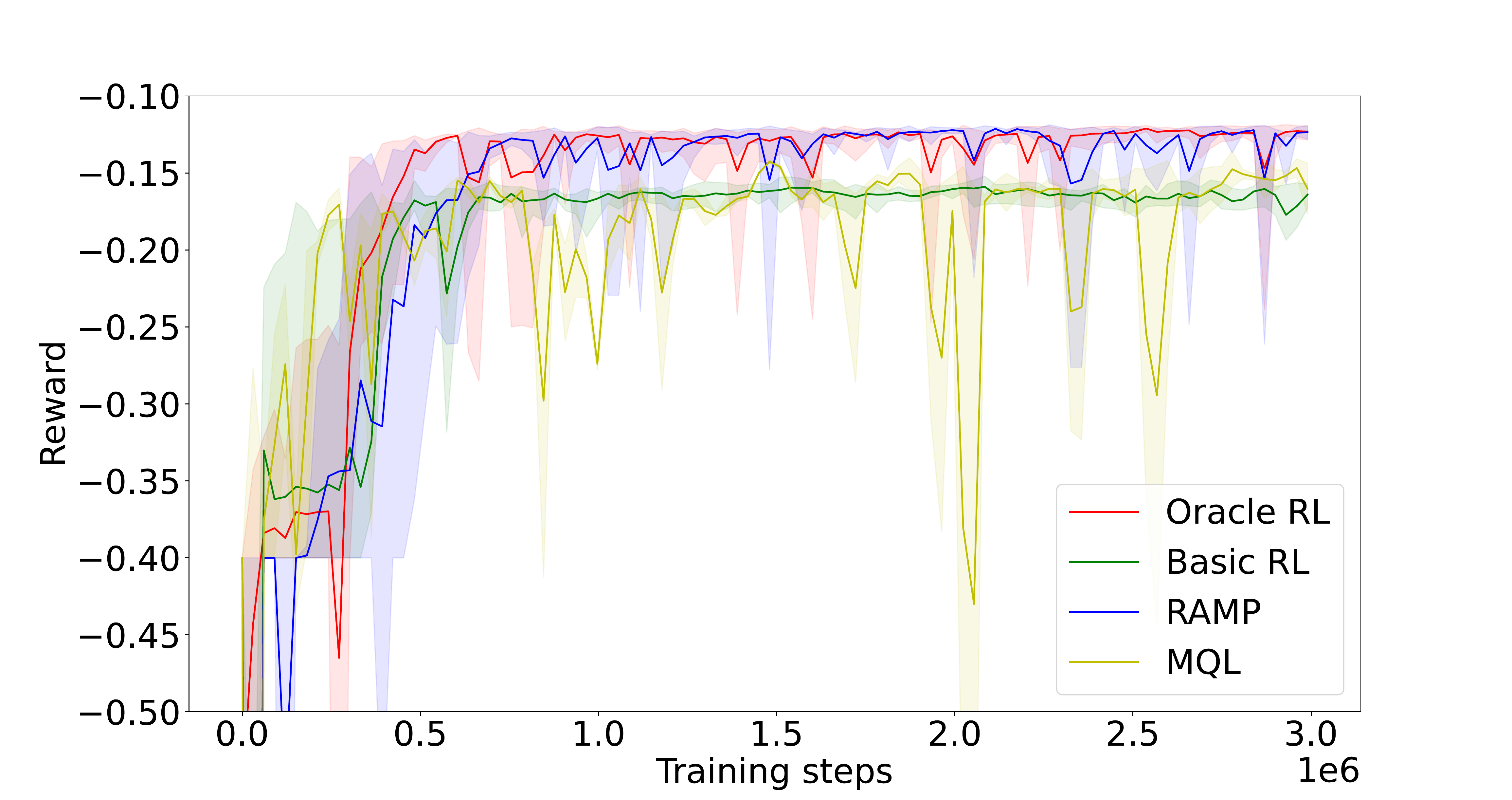}
    \caption{Comparison between the training processes. RAMP is close to the Oracle RL.}
    \label{fig:training_process}
\end{figure}

The training process of all methods was repeated $5$ times with different random seeds and is shown in Fig. \ref{fig:training_process}.
Our method's performance during the training process is comparable to the Oracle RL, achieving consistently higher episode rewards than the basic RL and MQL.  

Table \ref{Tab:performance} summarizes the average performance at the end of the training procedure. 
The table shows for all agents: the average reward in both test environments, the time to reach the target line by the vehicles with a low and high braking factor, and the average time.
As depicted by the table, RAMP reaches an average reward that is very close to the Oracle's and also has a very similar average time.
\begin{table}[h]
\centering
\begin{tabular}{|l|l|l|l|l|}
\hline
                     & \textbf{\begin{tabular}[c]{@{}l@{}}Average\\ Reward\end{tabular}} & \textbf{\begin{tabular}[c]{@{}l@{}}Low $\bm{k_a}$\\ Time {[}s{]}\end{tabular}} & \textbf{\begin{tabular}[c]{@{}l@{}}High $\bm{k_a}$\\ Time {[}s{]}\end{tabular}} & \textbf{\begin{tabular}[c]{@{}l@{}}Average\\ Time [s]\end{tabular}} \\ \hline
\textbf{Basic RL} & -0.1656       & 3.160                                                            & 3.376                                                                      & 3.268                                                  \\ \hline
\textbf{Oracle RL} & -0.1226       & 3.0                                                              & 1.976                                                                      & 2.488                                                  \\ \hline
\textbf{RAMP}        & -0.1230       & 3.048                                                            & 2.04                                                                       & 2.544                                                  \\ \hline
\textbf{MQL}         & -0.1510       & 3.18                                                              & 2.62                                                                      & 2.90                                                  \\ \hline
\end{tabular}
\caption{The performance of all RL agents on the test environments. The table shows for each agent: 1. The loss, averaged over the test episodes following the $5$ separate training processes; 2. The average time achieved by the vehicle with the low braking factor; 3. The average time achieved by the vehicle with the high braking factor; 4. The average between these times. }
\label{Tab:performance}
\end{table}

\begin{figure*}[h]
  \begin{subfigure}{0.5\textwidth}
      \includegraphics[width=\linewidth]{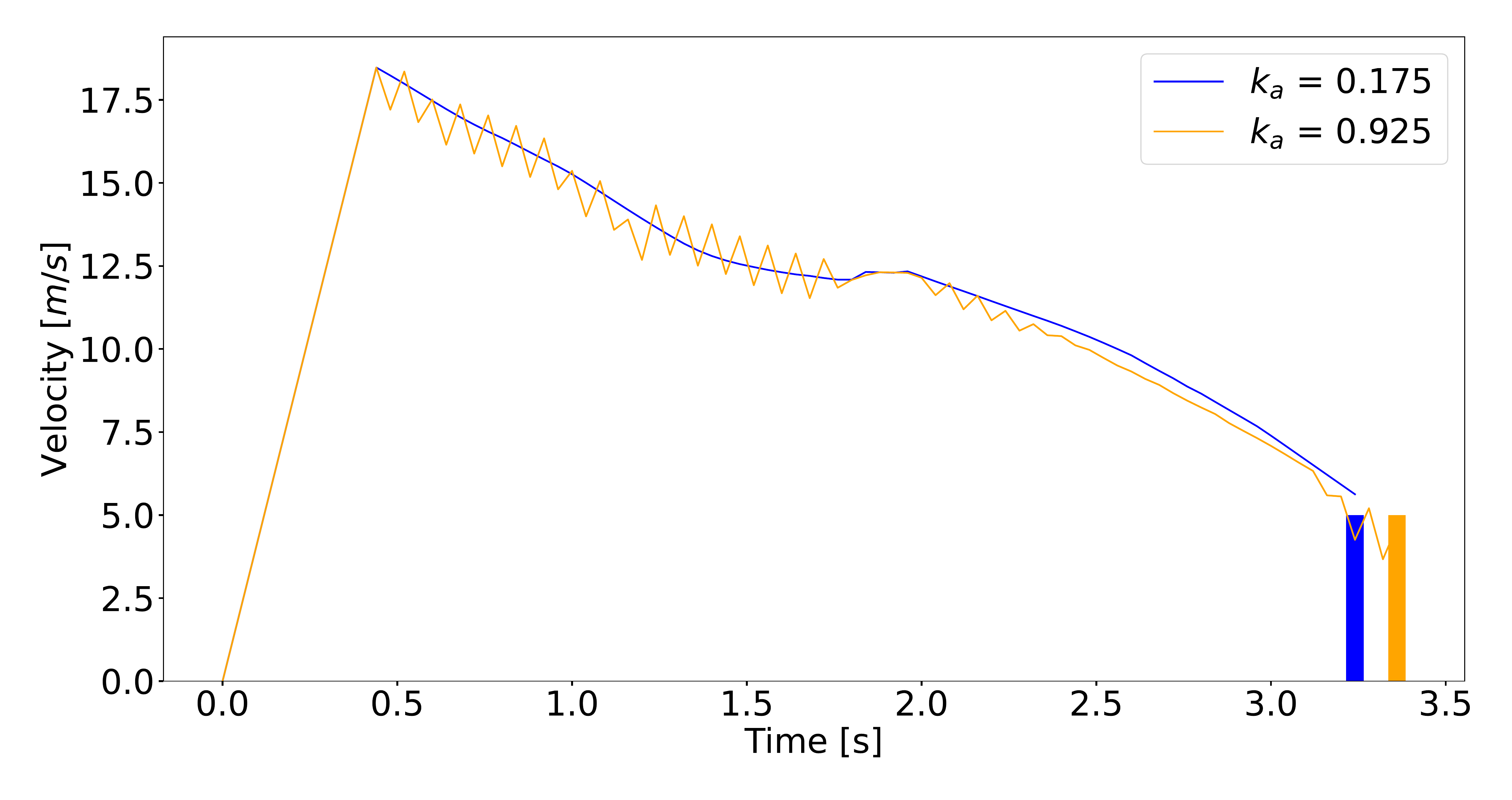}\hfill%
      \caption{}%
      \label{fig:trajectory_RL}%
  \end{subfigure}%
  \begin{subfigure}{0.5\textwidth}
      \includegraphics[width=\linewidth]{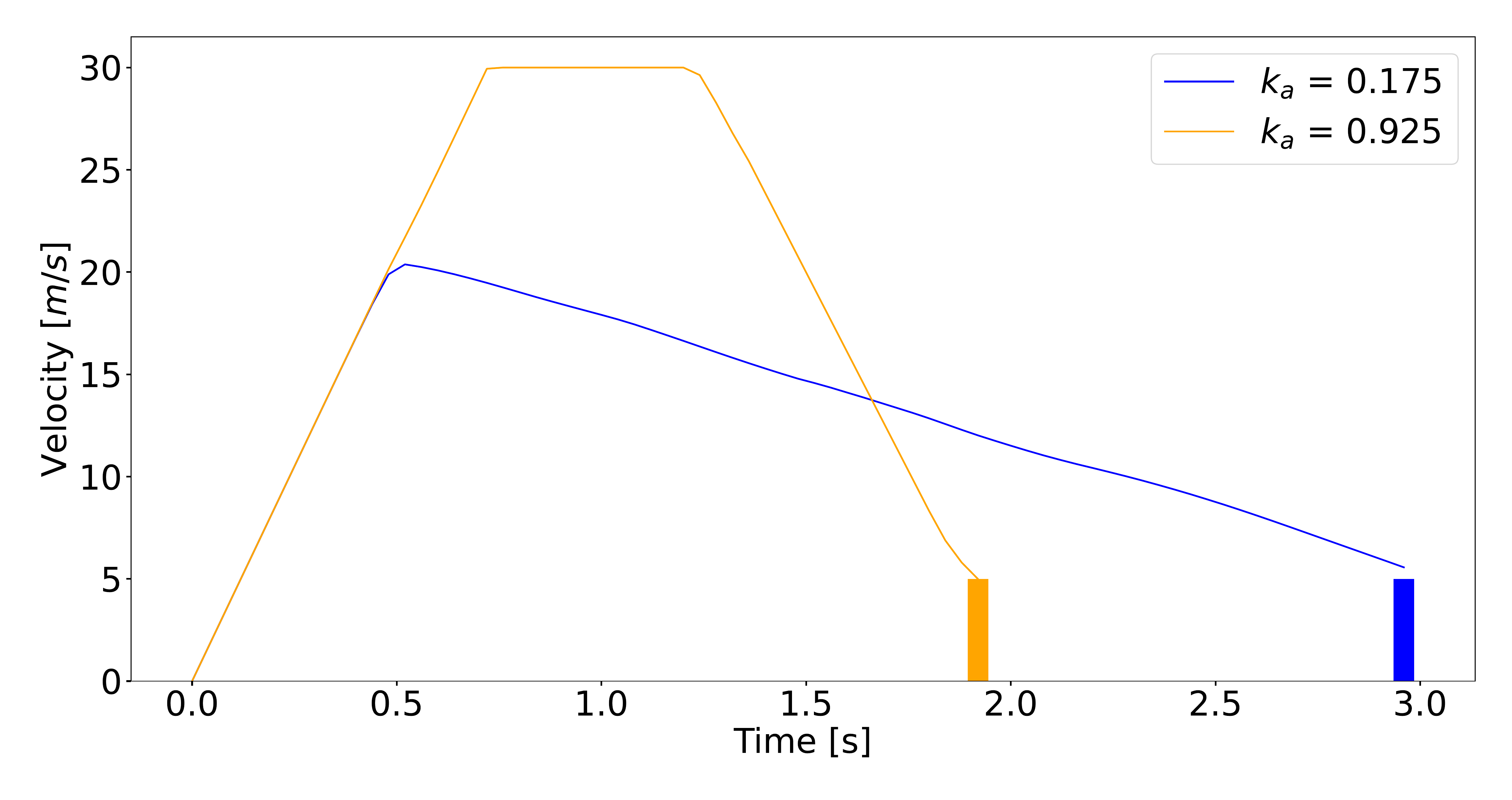}\hfill
      \caption{}
      \label{fig:trajectory_with_factor}
  \end{subfigure}\par\medskip
  
  \begin{subfigure}{0.5\textwidth}
      \includegraphics[width=\linewidth]{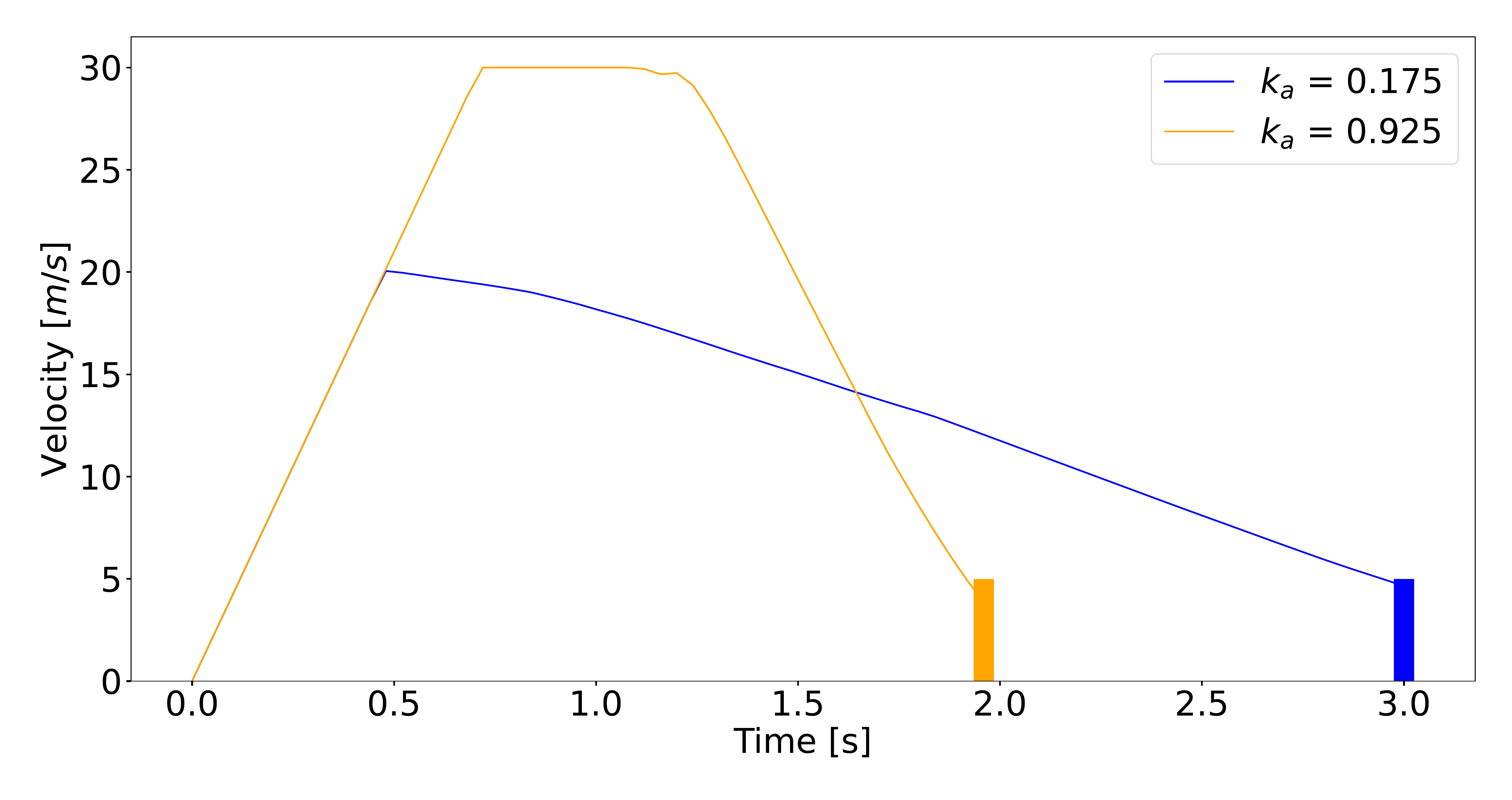}\hfill
      \caption{}%
      \label{fig:trajectory_RAMP}%
  \end{subfigure}%
  \begin{subfigure}{0.5\textwidth}
      \includegraphics[width=\linewidth]{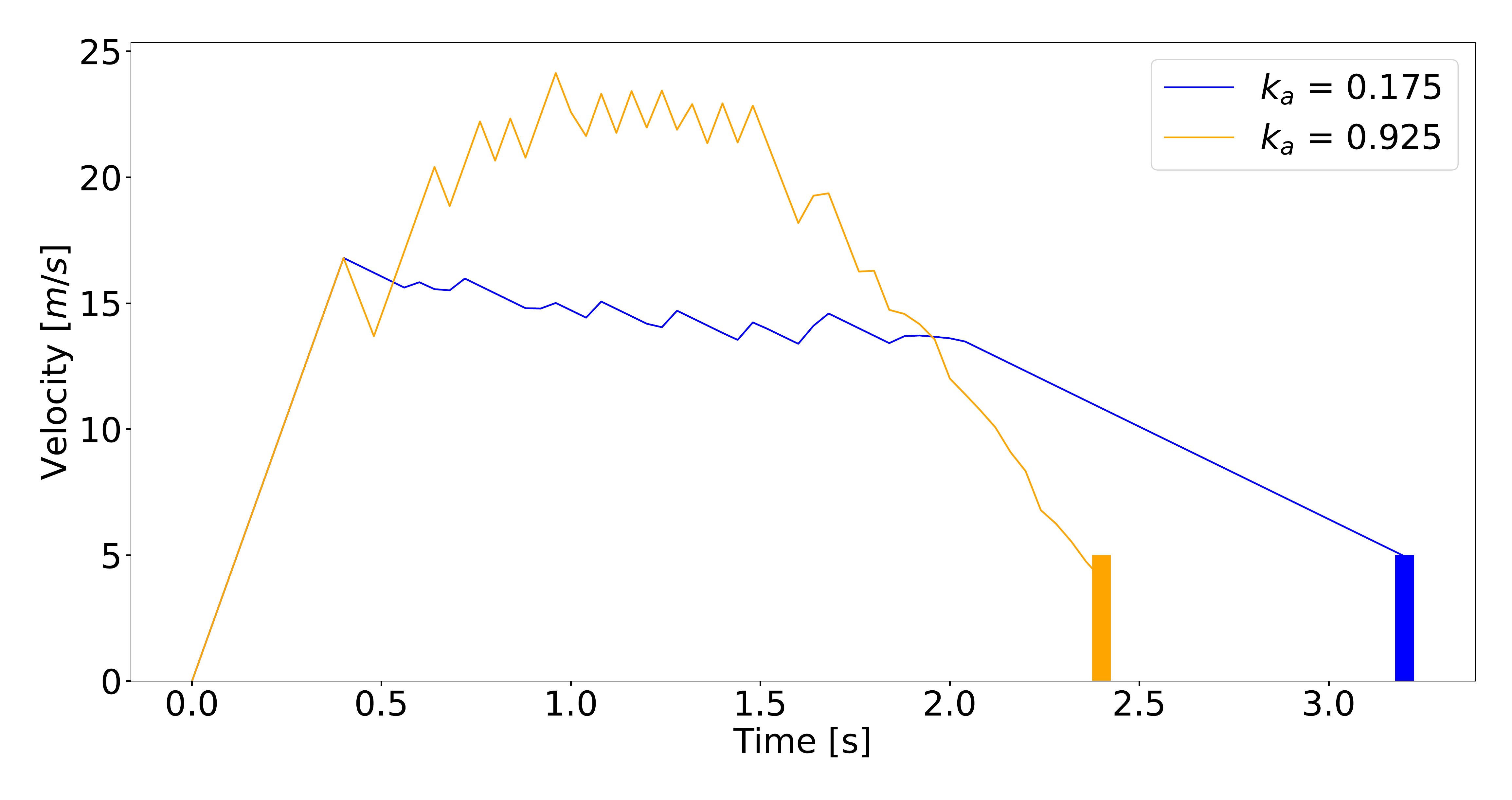}\hfill
      \caption{}
      \label{fig:trajectory_mql}
  \end{subfigure}\par\medskip
 
  \caption{Speed profiles that were achieved by our method and the other agents. Blue - low braking factor, orange - high braking factor, bold line - the maximal desired speed at the target. (a) The basic RL agent resulted in a conservative solution. (b) The Oracle RL agent achieved optimal speed profiles. (c) Our agent, RAMP, achieves similar optimal results without prior knowledge about the vehicle dynamics. (d) MQL resulted in sub-optimal results compared to RAMP.  }
  \label{fig:speed_profiles}
\end{figure*}

Next, we analyze the speed profiles of different vehicles driven by policies trained by the different RL agents. The speed profile of the basic RL, Oracle RL, RAMP, and MQL are shown in Figures \ref{fig:trajectory_RL}, \ref{fig:trajectory_with_factor}, \ref{fig:trajectory_RAMP}, and \ref{fig:trajectory_mql} respectively.
The orange lines represent speed profiles of vehicles with a high braking factor $k_a = 0.925$, and the blue lines represent low braking factors $k_a = 0.175$. These values are close to the extremes of the braking factor range to demonstrate the difference between the environments. The bold columns represent the maximal permitted speed at the target for each of the two environments. 
As depicted by Fig. \ref{fig:trajectory_RL}, the basic RL begins to brake on both vehicles at the same point in time. This happens because the agent cannot know if the vehicle has a higher braking capability that allows braking later or not, which leads to a conservative policy.
As shown in Fig. \ref{fig:trajectory_with_factor}, the Oracle RL begins braking on time in both environments and arrives at the destination at the required maximum target speed. 
As shown in Fig. \ref{fig:trajectory_RAMP}, RAMP results in a similar speed profile as the Oracle RL. However, unlike the Oracle agent, RAMP does not receive any explicit information about the environment; instead, it learns this information from the trajectory sampled during one episode.
Figure \ref{fig:trajectory_mql} illustrates that the MQL agent can distinguish between the vehicles' braking differences because the vehicle with the higher braking factor is allowed to gain more speed. However, MQL's speed profile is not as good as RAMP's since the MQL agent does not accelerate and decelerate at the maximal values, therefore resulting in longer driving times.

\begin{figure}[H]
    \centering
    \begin{subfigure}{0.8\linewidth}
        \centering
        \includegraphics[width=\linewidth]{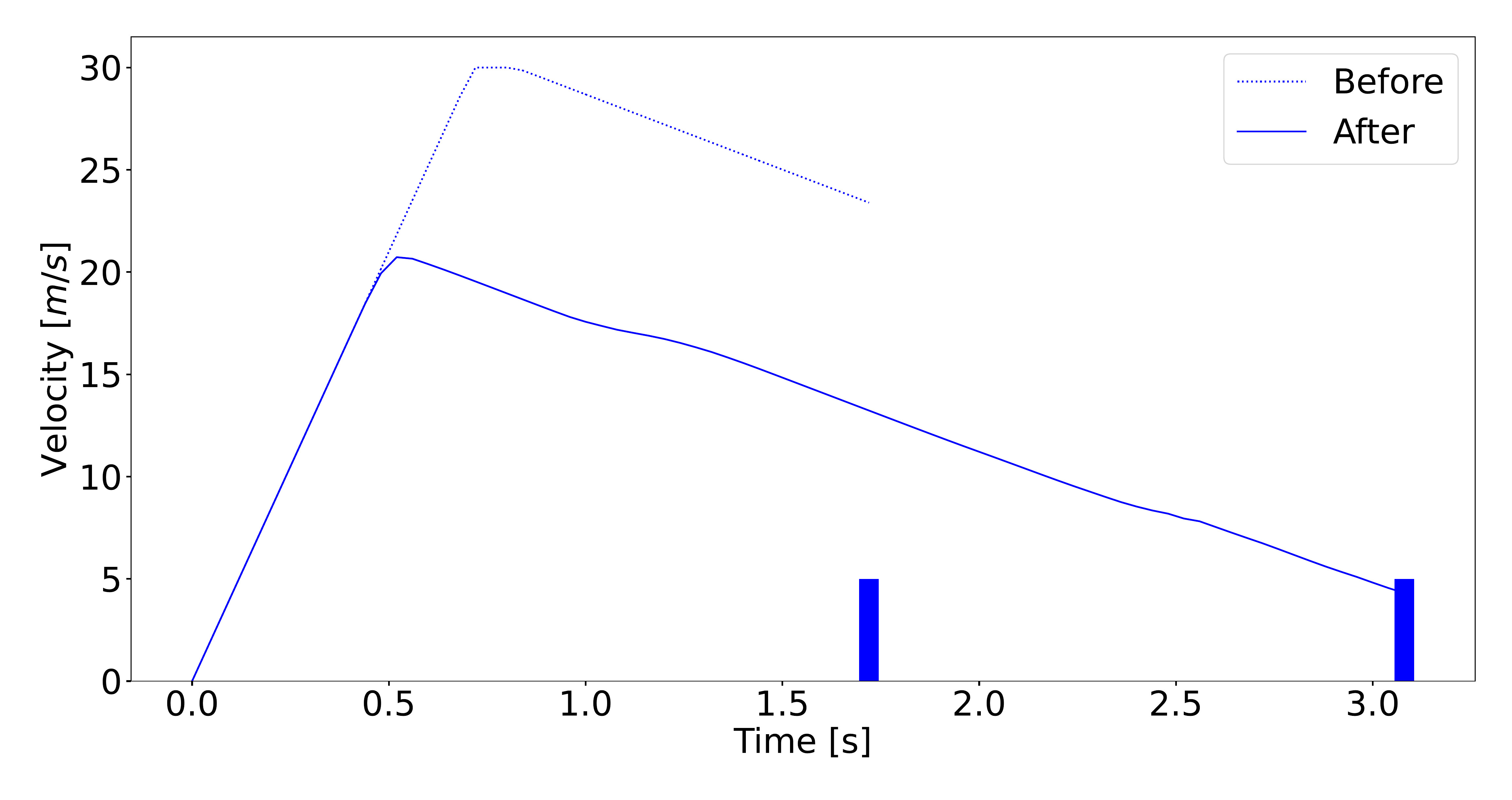}
        \caption{}
        \label{fig:trajectories_evaluation_low}
    \end{subfigure}
    \begin{subfigure}{0.8\linewidth}
        \centering
        \includegraphics[width=\linewidth]{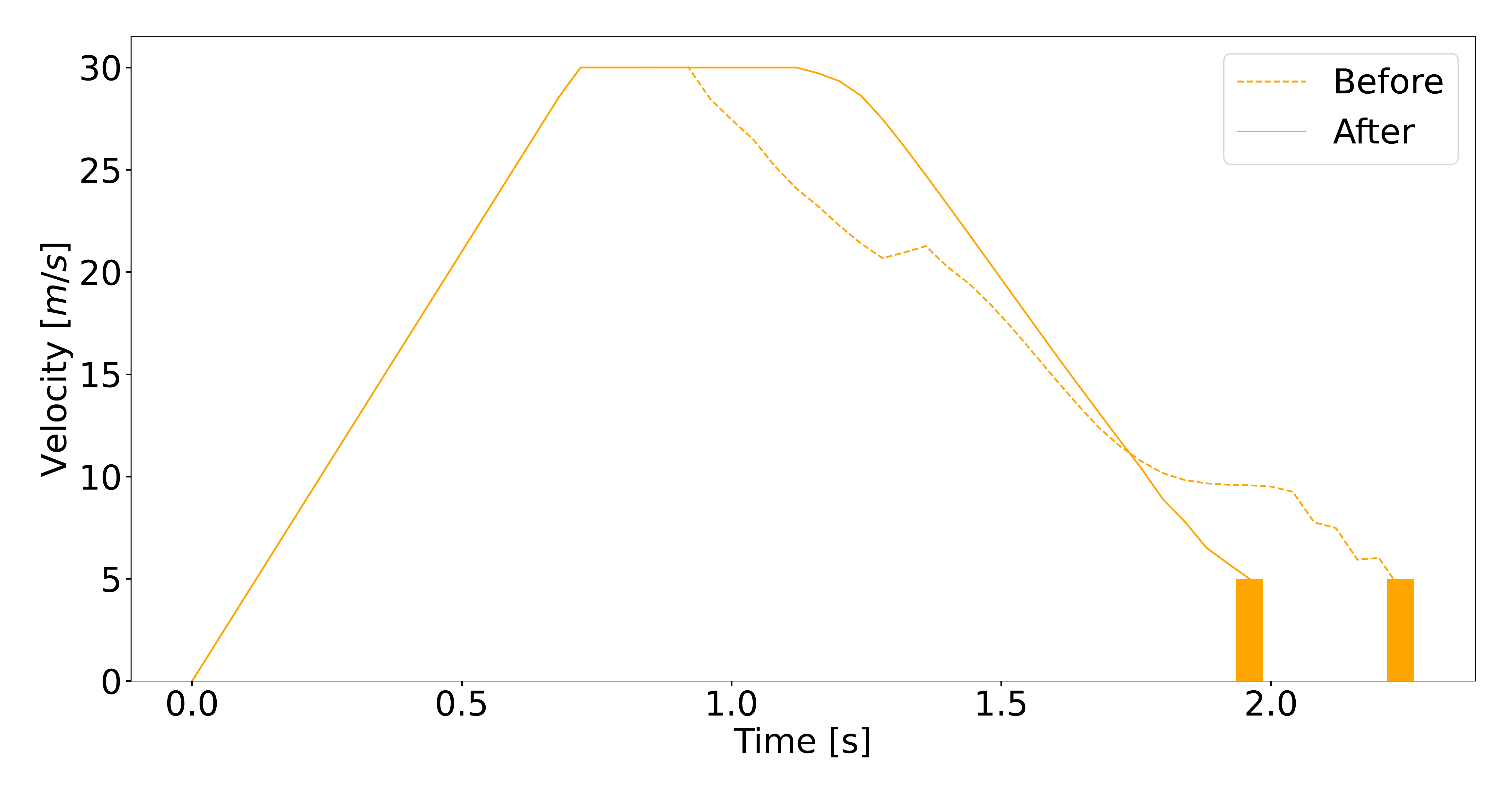}
        \caption{}
        \label{fig:trajectories_evaluation_high}
    \end{subfigure} 
\caption{RAMP evaluation during two subsequent episodes. The dashed line represents the speed profile used for collecting data in the first episode and the solid line in the second. (a) Low braking factor: the agent learned that this vehicle must brake earlier. (b) High braking factor: the agent learned that this vehicle can brake later.}
\label{fig:evaluation}
\end{figure}

To conclude the evaluation of RAMP's performance in the target-reaching domain, we analyze RAMP's adaptation process.
As opposed to the Oracle RL, which is given the braking factor information, RAMP must learn it from the driving experience. That is, in the first episode, RAMP collects data points, and the environment-specific parameters are trained on it; in the second episode, RAMP drives the vehicle with the updated context. Figure \ref{fig:evaluation} shows the speed profile of a vehicle during two subsequent episodes for the low braking factor vehicle (Fig. \ref{fig:trajectories_evaluation_low}) and for the high breaking-factor (Fig. \ref{fig:trajectories_evaluation_high}).
As depicted by Fig. \ref{fig:trajectories_evaluation_low}, in the first episode (represented by the dashed line), the vehicle brakes too late and therefore crosses the target line at too high a speed. In the second episode (represented by a solid line), the vehicle brakes earlier and cross the target line at a speed that is within the speed limit. 
Similarly, for the vehicle with a higher braking factor, RAMP learns that the vehicle can brake later. Therefore, in the second episode, it crosses the finish line earlier than in the first episode.

\section{Conclusions and Future Work}
This paper presented RAMP, a novel meta-reinforcement learning algorithm. RAMP is constructed in two phases:  learning a multi-environment dynamic model and training a general reinforcement learning policy that uses the model parameters as context.
The multi-environment dynamic model is trained on data from multiple environments. The shared parameters are updated by the average gradient computed from the loss resulting from all environments, and the environment-specific parameters are trained separately on data from each environment. The low number of environment-specific parameters allows direct use of them as context for the general policy. That general policy is trained by TD3, an actor-critic, off-policy RL algorithm.

We evaluated the performance of RAMP in simulated experiments. First, we tested the multi-environment dynamic model performance by a sine-wave regression test which we show to achieve a slightly lower loss compared to MAML \cite{maml}. Then, we tested RAMP in a simple driving domain where every vehicle had a different deceleration rate. 
We showed that RAMP achieved similar performance to an Oracle RL agent, which is provided with full knowledge of the environment properties.

In future work, we plan to test RAMP in more challenging domains, such as controlling the steering of an autonomous vehicle and following a given path.    %
Recall that RAMP assumes that the environments differ by their dynamics and not by their reward function, while most previous works consider the opposite. 
In order to adapt to environments that also differ by their reward functions, a future extension can be to learn a reward prediction function and uses its parameters as a context in addition to the multi-environment model parameters.

\bibliographystyle{IEEEtran.bst}
\bibliography{meta_rl_paper}

\end{document}